# Three-dimensional Human Tracking of a Mobile Robot by Fusion of Tracking Results of Two Cameras

Shinya Matsubara
*Department of Precision Mechanics*
*Chuo University*
Tokyo, Japan
matsubara@sensor.mech.chuo-u.ac.jp

Akihiko Honda
*Course of Precision Engineering*
Tokyo, Japan

Yonghoon Ji and Kazunori Umeda
*Department of Precision Mechanics*
*Chuo University*
Tokyo, Japan
{ji, umeda}@mech.chuo-u.ac.jp

*Abstract*— **This paper proposes a process that uses two cameras to obtain three-dimensional (3D) information of a target object for human tracking. Results of human detection and tracking from two cameras are integrated to obtain the 3D information. OpenPose is used for human detection. In the case of a general processing a stereo camera, a range image of the entire scene is acquired as precisely as possible, and then the range image is processed. However, there are problems such as incorrect matching and computational cost for the calibration process. A new stereo vision framework is proposed to cope with the problems. The effectiveness of the proposed framework and the method is verified through target-tracking experiments.**

## I. INTRODUCTION

Mobile robot systems are expected to be applied in many situations and environments - such as factories, construction sites, disaster scene and human living spaces [1]. In such environments, one of the important functions of a mobile robot system is to track a specific person. Studies of realizing such function have been actively conducted [2]-[6]. Stereo cameras are often used in these studies [7].

A general stereo camera obtains the range image of the entire image as accurately as possible. Then, processing according to the task is performed with the obtained range image. In order to fulfill such tasks, accurate calibration is necessary to obtain the range image accurately. However, there are problems such as incorrect matching and computational cost for the calibration process. In this paper, we propose a new stereo vision framework to obtain the optimal three-dimensional (3D) information to cope with the problems. We use OpenPose [8]-[10] to detect and track persons for individual cameras. By fusing the results, 3D tracking of persons is realized.

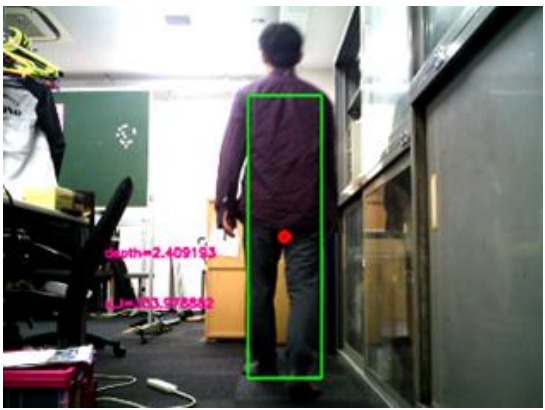

Fig. 1. 3D information about only required target.

## II. OVERVIEW

We propose a novel stereo vision framework to first perform processing according to a necessary task by each camera and then to fuse the processed results. By this framework, 3D information about only the necessary targets is obtained. In this paper, we focus on 3D human tracking as a specific task. After performing person detection and person tracked with each camera, 3D information of only the red point of the tracked person is acquired as shown in Fig. 1.

The overview of the proposed method is shown in Fig. 2. In our method, parallel stereo is realized by simply arranging two cameras of the same type in parallel on a mobile robot. Here, the cheap camera that is currently widely utilized is used to detect people. First, an image of the back view of the person being tracked is acquired, and a histogram is generated from the color information of the image, which is used as a template. Next, the human region in the camera image is detected using the OpenPose. Then we calculate the similarity between the histogram generated from the color information in the detected human region and the histogram of the template. If the similarity exceeds a threshold, it is determined to be a tracking target. Finally, the specified target is tracked based on OpenPose. At the same time, the distance to the specified target is measured, and the robot is controlled based on the calculated distance value. In other words, the tracked person is specified from the information on the color of the clothes in this method.

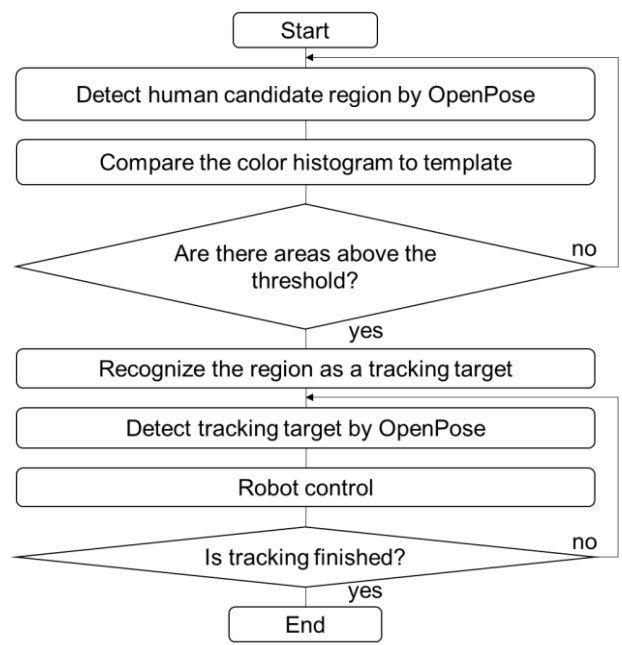

Fig. 2. Flow chart.

## III. PERSON TEMPLATE GENERATION

In this study, the acquired RGB image is converted to HSV color space and used to generate a color information template. The HSV color space represents hue, saturation, and lightness, respectively. Next, a histogram of the template is generated using only hue information that is not easily affected by the brightness of the surrounding environment or lighting conditions. In this method, the part below the shoulder of the person is used as a template. Figure. 3 shows an example of a template and its hue histogram.

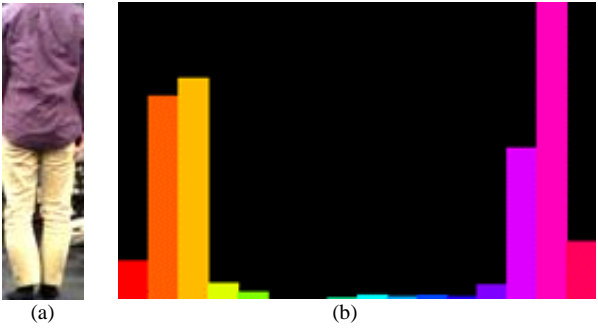

(a)　　　　　　　　　(b)

Fig. 3. Histogram generation for template image: (a) template image and (b) color histogram.

## IV. MOBILE ROBOT TRACKING

The tracked human region is detected by the left and right cameras by the above method. The disparity is calculated from the difference between the center points of the human region detected by each of the cameras as shown in Fig. 5. From that disparity of the two cameras, the distance to the target is calculated as follows:

$$Z = \frac{b \cdot f}{\delta(u_l - u_r)} \quad (2)$$

where, $u_l$ and $u_r$ are the abscissas of the center points of the left and right cameras; thus, $u_l - u_r$ indicates the disparity of the each of cameras. $b$ is the length of the base line between each of cameras. $f$ is the focal length of the camera and $\delta$ is the pixel interval.

Using the detection result of the tracked target person and the distance information, the mobile robot is controlled in online processing and tracks a specific person. PID control is used to control the mobile robot. The control of the speed of the mobile robot is set as the distance and direction between the human and the mobile robot.

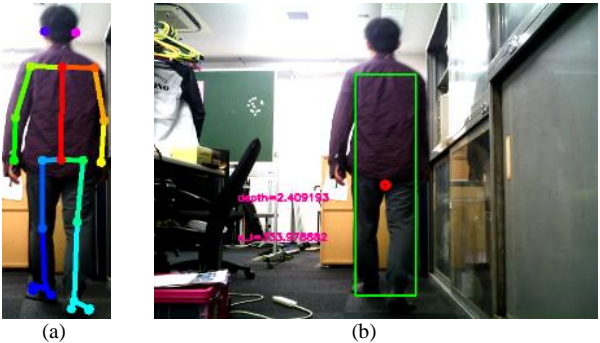

(a)　　　　　　　　　(b)

Fig. 4. Detection of human region: (a) detection of human body key points by OpenPose and (b) examples of region used to calculate histogram similarity.

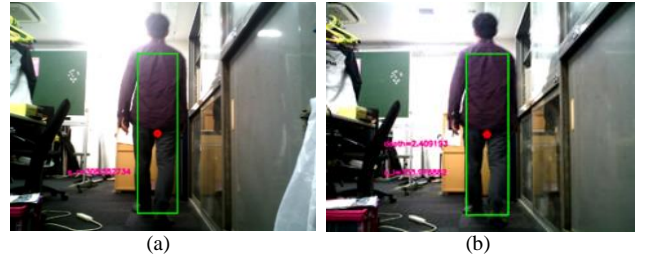

(a)　　　　　　　　　(b)

Fig. 5. Examples of points used for distance calculation: (a) left camera image and (b) right camera image.

## V. EXPERIMENT

### A. Overview of tracking experiment

An experiment was performed to verify the effectiveness of the proposed method. An additional experiment was also performed in an environment where tracking was impossible with the conventional method. In the conventional method, when the lighting conditions change, it becomes impossible to correctly recognize the person to be tracked, and tracking fails. Therefore, we conducted experiments in an environment where lighting conditions change. In addition, we verified whether tracking can be performed even when another person crosses between the tracked person and the mobile robot as shown in Fig. 10.

The experiments ware performed indoors in two places. One was in a relatively bright environment where the light conditions change, and the other was a relatively dark environment where the light conditions change. We performed human tracking using two web cameras and a mobile robot. The web camera was ELECOM UCAM-DLI500TN as shown in Fig. 6 (a). The two cameras were used side by side. The image resolution was 640 × 480 pixels and the angle of view was 54 °. The length of a base line was 94 mm. The mobile robot used in our experiments was Pioneer 3-AT (MobileRobots) as shown in Fig. 6 (b). When the distance between the person and the mobile robot exceeds a certain value, the robot starts moving toward the person, as shown in Fig. 7.

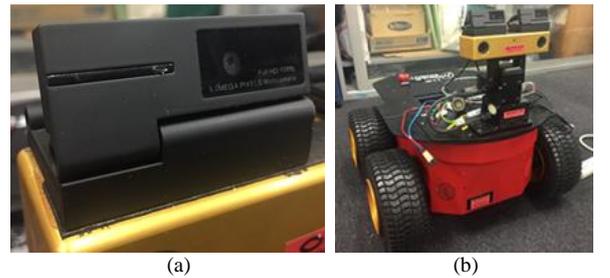

(a)　　　　　　　　　(b)

Fig. 6. Experimental equipment: (a) web camera and (b) mobile robot.

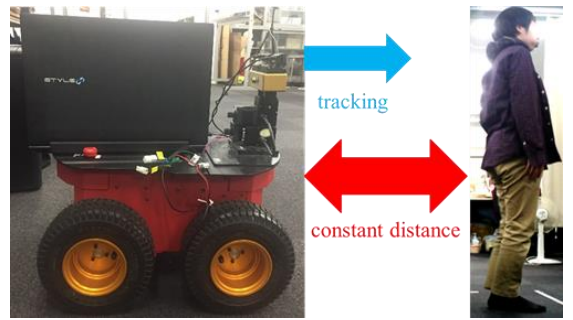

Fig. 7. Conceptual diagram of target tracking with a mobile robot.

## B. Experimental results

Figures 8 and 9 show the experiments performed in the relatively bright and relatively dark environments, respectively. Fig. 10 shows the experiment in which a person crossed between the tracked person and the mobile robot. The experiment in Figure 10 was performed in the same place as the experiment in Fig. 9.

The robot was set to move forward when the robot was 2 m away from the target person. Human tracking was successful in both experimental environments. However, the robot sometimes stopped in the relatively dark environment as shown in Fig. 9. Because the speed at which the person walks and the speed at which the mobile robot moves are different, the distance between the person and the robot is not constant. Nevertheless, tracking was successfully performed.

Figure 11 shows the distance data between the tracked person and the robot in the experiment shown in Fig. 10 The vertical axis represents distance, and the horizontal axis represents time. The red box indicates the periods when the person was not detected. Fig. 11 shows the mobile robot was able to track a person in an environment where occlusion occurred. However, the distance between the tracked person and the mobile robot varies slightly.

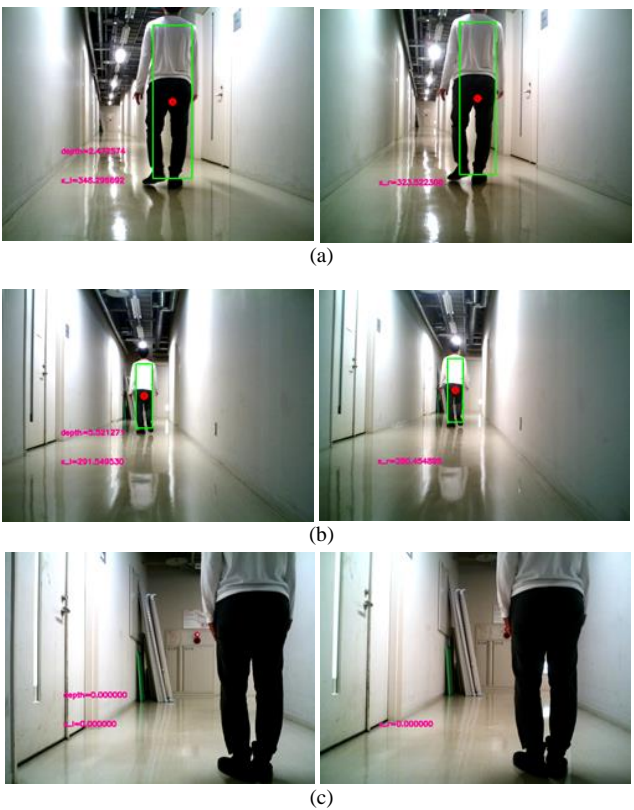

Fig. 8. Experimental results of two cameras in relatively bright indoor environment: (a) just before robot moves, (b) while moving, and (c) immediately after robot stops.

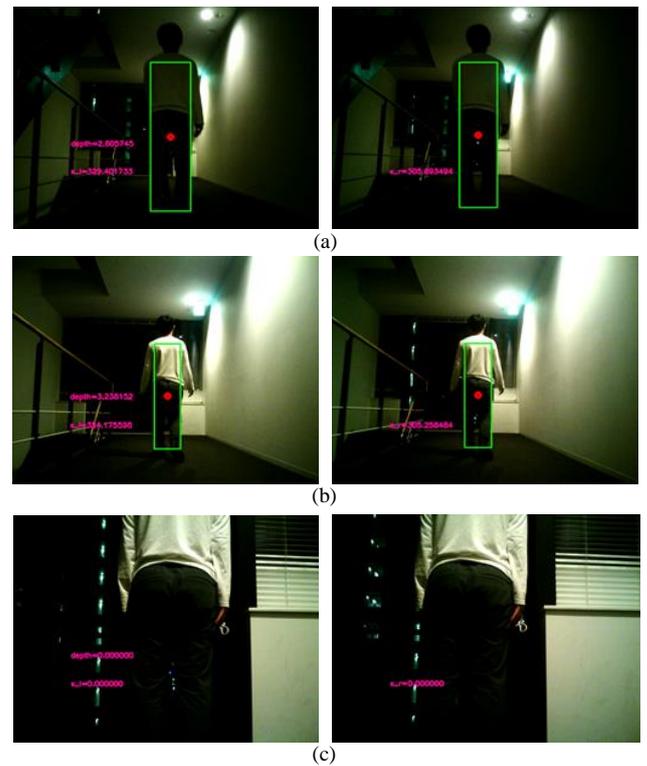

Fig. 9. Experimental results of two cameras in relatively dark indoor environment: (a) just before robot moves, (b) while moving, and (c) immediately after robot stops.

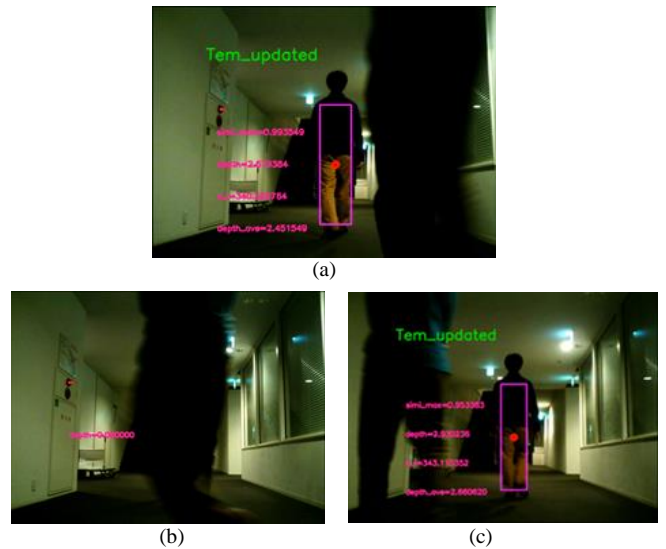

Fig. 10. Experimental results of scenes where person cross: (a) before person crossed, (b) while person were crossing, and (c) after person crossed.

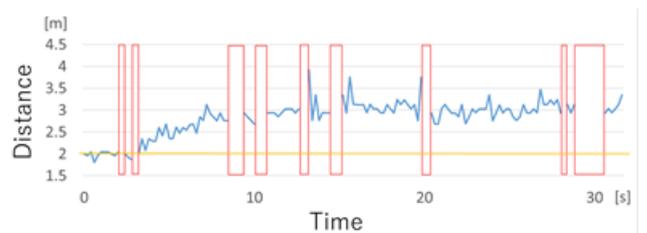

Fig. 11. Time lapse and distance between tracked person and mobile robot.

*C. Discussion*

In the first experiment shown in Fig. 8, although the environment was bright, the tracked person may look bright or dark due to the influence of lighting on the way. Figure 8 (a) shows the situation at the start of tracking. At this time, the tracked person looks dark from the perspective of the robot. On the other hand, Fig. 8 (b) shows the situation when a certain time has passed after the robot started tracking and the person moved forward. At this time, the person being tracked looks bright when viewed from the robot. This experiment was conducted in a place where such lighting fluctuations occur many times. Even in the case of such lighting fluctuations, the mobile robot was able to track the person when the person moved forward and the mobile robot kept a certain distance away from the person.

In the second experiment shown in Fig. 9, as in the first experiment, the person being tracked may look bright or dark due to the influence lighting condition on the way. In addition, the experiment was performed in a darker environment than in the first experiment. Figure 9 (a) shows the situation at the start of tracking. At this time, the human being tracked looks darker than the first experiment. On the other hand, Fig. 9 (b) shows the situation when a certain time has passed since the robot started tracking and the person moved forward. At this time, the person looks bright when viewed from the robot due to the effects of lighting in a dark environment. This experiment was conducted in a place where such lighting fluctuations occur repeatedly. The mobile robot was able to track the person even if such lighting fluctuations occurred. Moreover, in both experiments, it was possible to track a person who stopped once and started walking again. As shown in Figs. 8 (c) and 9 (c), the whole person was not captured by the camera at the point where the robot stopped in both experiments. This is because when the person stops, the robot moves forward between the time the robot is determined to stop and the actual stop of the robot, and the robot approach the person.

The robot sometimes stopped in the relatively dark environment shown in Fig. 9. This is probably because the surrounding environment was too dark to detect people with OpenPose.

## VI. CONCLUSION

In this paper, we proposed a new stereo vision framework that obtained only the 3D information needed to track a specific person by fusing the processing results of the two cameras. Moreover, we proposed a new human tracking system. We used OpenPose to track a specific people. As a result, we showed that a mobile robot can track a person in an environment where it was impossible to track with the conventional method.

In this experiment, it is considered that the environment was easy to track a person because it was performed indoors and there were few persons who appear in the camera. Therefore, it is necessary to verify whether it can be tracked outdoors or in an environment where many persons exist. Moreover, in the current method, the person is identified only by color and tracked; thus, if the color changes due to a sudden change of lighting conditions, it is difficult to track. It is also difficult to track in an environment with multiple people wearing the same color clothes. In order to solve these problems, it is necessary to use other features in addition to color information.